\documentclass[11pt]{article}
\usepackage[]{acl}

\usepackage[colorinlistoftodos]{todonotes}

\usepackage{times}
\usepackage{latexsym}

\ifxetex
    \usepackage{fontspec}
\else
    \usepackage[T1]{fontenc}
    \usepackage[utf8]{inputenc}
\fi

\usepackage{url}

\usepackage{graphicx} 
\graphicspath{{images/}}

\usepackage[inline]{enumitem}
\usepackage{bbm}
\usepackage{verbatim}
\usepackage{multirow}
\usepackage{tabularx}
\usepackage{booktabs}
\usepackage{microtype}
\usepackage{pbox}

\usepackage{booktabs}

\usepackage[subtle]{savetrees}

\usepackage{amsfonts}
\usepackage{amssymb}
\usepackage{amsthm}
\usepackage{amsmath}

\usepackage{setspace}

\usepackage{xspace}

\makeatletter
\newcommand*{\etc}{%
	\@ifnextchar{.}%
	{etc}%
	{etc.\@\xspace}%
}
\makeatother

\usepackage{xcolor}

\definecolor{nice-red}{HTML}{E41A1C}
\definecolor{nice-orange}{HTML}{FF7F00}
\definecolor{nice-yellow}{HTML}{FFC020}
\definecolor{nice-green}{HTML}{4DAF4A}
\definecolor{nice-blue}{HTML}{377EB8}
\definecolor{nice-purple}{HTML}{984EA3}

\newcommand{\opAnd}{\operatorname{AND}}
\newcommand{\opOr}{\operatorname{OR}}
\newcommand{\opNot}{\operatorname{NOT}}

\usepackage{todonotes}
\makeatletter
\newcommand*\iftodonotes{\if@todonotes@disabled\expandafter\@secondoftwo\else\expandafter\@firstoftwo\fi}
\makeatother

\title{
EVI: Multilingual Spoken Dialogue Tasks and Dataset for \\
Knowledge-Based Enrolment, Verification, and Identification
}

\author{
    \vspace{2mm}
    \textbf{Georgios P. Spithourakis},
    \textbf{Ivan Vuli\'{c}}, \\
    \vspace{2mm}
    \textbf{Micha\l{} Lis},
    \textbf{I\~{n}igo Casanueva},
    {\normalfont and} \textbf{Pawe\l{} Budzianowski} \\
    \vspace{2mm}
    PolyAI Limited \\
    London, United Kingdom \\
    \texttt{\small\{georgios,ivan,michal,inigo,pawel\}@poly.ai} \\
}

\newcommand{\sparagraph}[1]{\noindent\textbf{#1.}}
\newcommand{\rparagraph}[1]{\vspace{1.4mm}\noindent\textbf{#1.}}

\date{}

\begin{document}
\maketitle




\begin{abstract}
Knowledge-based authentication is crucial for task-oriented spoken dialogue systems that offer personalised and privacy-focused services. Such systems should be able to \textit{enrol} (E), \textit{verify} (V), and \textit{identify} (I) new and recurring users based on their personal information, e.g. postcode, name, and date-of-birth. In this work, we formalise the three authentication tasks and their evaluation protocols, and we present~\textit{EVI}, a challenging spoken multilingual dataset with $5,506$ dialogues in English, Polish, and French. Our proposed models set the first competitive benchmarks, explore the challenges of multilingual natural language processing of spoken dialogue, and set directions for future research.

\end{abstract}

\section{Introduction}
\label{sec:intro}
Computer systems need to be able to identify and verify their users
before granting access to personalised services and confidential information~\cite{braz2006security,o2003comparing}.
In particular,
\textbf{identification (I)} is the process of
specifying the identity of a person,
i.e. answer the question: \textit{``who are you?''}.
On the other hand,
\textbf{verification (V)} (aka authentication) is the process of
confirming the assertion about a claimed identity,
i.e. answer \textit{``are you who you claim you are?''}~\cite{jain2004introduction}.
In both processes,
the system compares information given by the user
with information held by the system;
thus they presume \textbf{enrolment (E)},
that is, the process of registering the identity information of a new user into the system~\cite{jain2004introduction}.

\begin{figure}[!t]
\centering
\includegraphics[width=0.8\columnwidth]{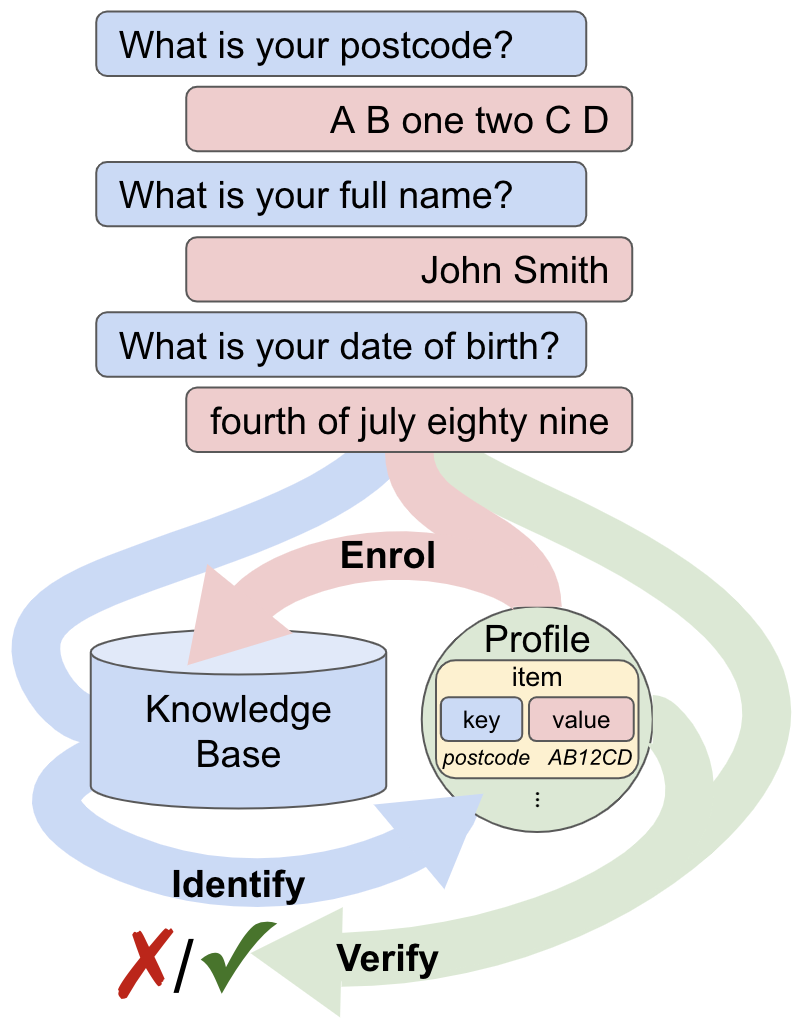}
\caption{
Knowledge-based EVI for task-oriented spoken dialogue systems:
enrolment (E) creates a new user profile to store in a KB;
identification (I) retrieves a pre-enrolled profile for a user;
and verification (V) asserts whether the user matches a claimed profile.
}
\label{fig:tasks}
\end{figure}

Task-oriented dialogue systems
that offer personalised and privacy-focused services (e.g. set up utilities, track a parcel, or access a bank account) should be able to enrol, identify, and verify new and recurring users, without interrupting their natural conversational interface.
Different types of authentication factors may be used~\cite{smith2001authentication,o2003comparing}:
\begin{enumerate*}[label=\roman*)]
  \item knowledge-based ("\textit{what you know}"),
  rely on a secret \textit{password} or personal information,
  e.g. full name, date of birth, mother's maiden name, etc.;
  \item possession-based ("\textit{what you have}"),
  rely on possession of a physical \textit{token},
  e.g. a smart card, a metal key, etc.; and
  \item inherence-based ("\textit{who you are}"),
  typically rely on \textit{biometric} properties,
  e.g., a voiceprint, fingerprint, eye scan, or signature \cite{variani2014deep}.
\end{enumerate*}
Most businesses use knowledge-based authentication in their call centres to identify customers over the phone~\cite{hrabi2020call,amein2020hidden,morgen2012voice,petersen2019complexity}. As conversational AI is increasingly being used to automate call centres,
we seek to enable task-oriented spoken dialogue systems with EVI functionalities.

The core \textbf{contributions} of this paper are:
\begin{enumerate}[topsep=1pt,itemsep=-1ex,partopsep=1ex,parsep=1ex]
   \item We motivate and formalise knowledge-based enrolment, verification, and identification as \textbf{novel tasks} for task-oriented spoken dialogue systems (Section~\ref{sec:tasks}).
   \item We collect and publish a \textbf{novel conversational dataset} with $5,506$ dialogues that can be used to develop and evaluate EVI-oriented spoken dialogue systems in $3$ languages (British English, Polish, and French; Section~\ref{sec:data}). The multilingual aspect of the dataset allows us to also study language-specific variations in data and performance, reaching beyond monolingual, English-only setups.
   \item {We define baseline models and suitable evaluation protocols (Section~\ref{sec:methods}) for the new tasks.
   Finally, we evaluate \textbf{benchmarks} on the new dataset,
   explore the unique challenges of these tasks,
   and set directions for future research (Section~\ref{sec:experiments})}.
\end{enumerate}
The code and dataset is available online at: {\url{https://github.com/PolyAI-LDN/evi-paper}.

\section{The EVI Dialogue Tasks}
\label{sec:tasks}


\sparagraph{Preliminaries}
For all tasks,
we assume that the dialogue system can interact with a \textit{Knowledge Base (KB)} of stored profiles, $P_\textit{KB}=\{p_1, p_2, ...\}$.
Each \textit{profile}, $p$, is a structured record
of a real-world entity (e.g. a user, product, etc.)
that comprises one or more \textit{items},
i.e. key-value pairs
(e.g. postcode, name, date of birth, etc.).
The user and system take alternate turns, $t$, that make up a multi-turn \textit{dialogue},
$T_\textit{dialogue}=\{t_\textit{1,system},t_\textit{1,user},t_\textit{2,system},t_\textit{2,user},...\}$.

\rparagraph{Enrolment Task}
The goal of enrolment is to
create and store a profile
that represents the identity of a \textit{new user}
and that can be used to identify or verify the same user
in the future.
For dialogue-based enrolment,
the system must be able to
extract all required item key-value pairs from the dialogue
to construct a new profile to store in the KB (cf. Fig.~\ref{fig:tasks}):
\begin{equation}
p_\textit{new}={\textrm{enrol}}(T_\textit{dialogue})
\end{equation}
\sparagraph{Verification Task}\,The goal of verification is to
decide whether a user who claims an identity
is \textit{genuine} or an \textit{impostor}.\,For dialogue-based, knowledge-based verification,
the system must be able to
compare information stored in the KB about the claimed identity
with information provided by the user in the dialogue
to produce a verification \textit{score}
that quantifies the degree of the match
(cf. \!Fig.~\!\ref{fig:tasks}):
\begin{equation}
s_\textit{profile} = {\textrm{verify}}(p_\textit{claimed}, T_{dialogue}) 
\in [0,1],
\label{eq:verify-profile}
\end{equation}
where $s=1$ signifies a genuine verification attempt,
and $s=0$ denotes an impostor verification attempt.
The system designer can apply a threshold, $\theta$,
to obtain a crisp verification outcome
and control the system's trade-off between security and usability (see later Subsections~\ref{sec:verification-policy} and~\ref{sec:evaluation}).


\rparagraph{Identification Task}
The goal of identification is to
determine the identity of an unknown user
from a KB of pre-enrolled user profiles.
For dialogue-based, knowledge-base identification,
the system must be able to query the KB with the information provided by the user in the dialogue to retrieve a ranked list of the best matching profiles
(cf. Fig.~\ref{fig:tasks}):
\begin{equation}
p_{1}, p_{2}, ...  = {\textrm{identify}}(P_\textit{KB}, T_\textit{dialogue})
\end{equation}
The list might be empty if no qualifying profiles (i.e. above a score threshold) could be retrieved.

\section{{\fontdimen2\font=0.4ex{A~Multilingual~Spoken~Dialogue~Dataset}}}
\label{sec:data}

We set out to build a novel, first of its kind, human-to-machine conversational dataset that can be used to develop and evaluate task-oriented spoken dialogue systems that support the functionality of the knowledge-based EVI tasks. The dataset is multilingual and covers $3$ locales: British English (en-GB), French (fr-FR), and Polish (pl-PL).\footnote{The choice of these languages was motivated by the popularity, the phonetic richness and a large enough base of high-quality crowdworkers.}

\subsection{Generating the Profiles Knowledge Base} \label{sec:kb-generation}

For each locale, we populate a KB to be shared across EVI tasks.
We randomly generated locale-dependent profiles using the \textit{faker} tool.\footnote{\url{https://faker.readthedocs.io/}; it is a python package that can generate fake but reasonable data (names, addresses, phone numbers, etc.) for bootstrapping databases.} Each profile in the KB consists of its generated item key-value pairs for \textit{postcode}, \textit{full name}, and \textit{date of birth} (cf. Fig.~\ref{fig:tasks}). These three different slots are popular in industrial authentication procedures. Because in the real world people might share the same
name, postcode, or date of birth by coincidence,
we allow duplicate values in our generated data,
e.g. for each locale our KB contains $10,000$ unique profiles,
but only $2,000$ unique postcodes. Table~\ref{tab:describe_data} shows the size of the generated KB.

\subsection{Collecting the Dialogue Data}
\label{sec:data-collection}

We developed a \textbf{spoken dialogue system}
to collect the postcode, full name, and date of birth
of a user over the phone.
The system operates under a deterministic policy
with static retries for each collection step.
We use the same sequence of dialogue acts for all EVI tasks,
and vary the scripted prompts (see Subsection~\ref{sec:data-analysis}) to elicit more diverse responses:
\begin{quote}
{\small
\begin{enumerate}[label=Q\arabic*:,align=left, leftmargin=*]
    \item What is your postcode?
    \item Please tell me your postcode.
    \item I heard [A B 1]. Please tell me your postcode.
    \item What is your full name?
    \item Please tell me your first and last name.
    \item Please spell your full name.
    \item What is your date of birth?
    \item Please tell me your date of birth.
    \item I heard [the 1st of January]. Please tell me your date of birth.
\end{enumerate}}%
\end{quote}

For other locales, see Appendix~\ref{sec:appendix-qs}. For each locale, we enlisted cohorts of speakers on the \emph{Prolific Academic} (\url{www.prolific.co}) crowdsourcing platform. We displayed a random profile from the KB for each speaker to impersonate, e.g.:

\vspace{-2.5mm}
{\small
\begin{align*}
\textbf{Postcode}&: \textit{AB1 2CD} & \textbf{Kod Pocztowy}&: \textit{12-345} \\
\textbf{Full Name}&: \textit{John Smith} & \textbf{Imię i Nazwisko}&: \textit{Anna Krupa} \\
\textbf{Date of Birth}&: \textit{4/7/1989} & \textbf{Data urodzenia}&: \textit{1/1/2000}
\end{align*}}%
Then, we directed speakers to call a phone number to interact with our spoken dialogue system. To ensure quality, the crowdsourced speakers had to complete all turns of the static policy to receive their payment code.\footnote{The workers were not aware that the system was scripted, yielding the natural behaviour of irritated customers.}
Additionally, we filtered out all dialogues for which text-to-speech detected silence for all turns of a single item or for more than half of the turns of the dialogue.

For each turn, the EVI conversational dataset contains:
the unique identifier of the impersonated profile from the KB;
a unique speaker identifier;
the raw audio data;
the n-best list of transcriptions (see Appendix~\ref{appendix-asr});
and any variation in the prompts
(see Subsection~\ref{sec:data-analysis}).
Table~\ref{tab:describe_data} shows the size of our dialogue dataset for all locales, which contains $5,506$ dialogues in total.

\subsection{{Diversifying Speaker Behaviours}}
\label{sec:data-analysis}

{
To elicit mode diverse behaviours from the speakers,
and thus increase the generality and richness of our dataset,
we exploited two psychological phenomena: priming and entrainment.
}

\textit{Priming} is the psychological effect 
wherein exposure to a stimulus (\textit{prime})
unconsciously influences the response to a later stimulus (\textit{target}).
Priming also affects linguistic decision making,
e.g.
exposure to a lexical item or syntactic structure
reinforces reuse of the same pattern in the future~\cite{reitter2006computational,reitter2010priming}.
Likewise,
\textit{entrainment} is the phenomenon wherein
conversational interlocutors adopt each other's linguistic patterns.
Entrainment can be observed at multiple levels,
e.g.
lexical~\cite{brennan1996conceptual},
syntactic~\cite{reitter2007predicting},
stylistic~\cite{niederhoffer2002linguistic},
phonetic~\cite{pardo2006phonetic},
and prosodic~\cite{coulston2002amplitude}.
The Interactive Alignment Model~\cite{pickering2004toward}
proposes that conversational interlocutors automatically prime each other at multiple levels, causing their speech to converge.\footnote{Alternatively, Communication Accommodation Theory~\cite{giles19911}
proposes that more strategic decisions drive convergence (or divergence).}

\rparagraph{Diversifying Spoken Dates} Our primes
to diversify the speakers' lexical choice for dates
were the formats that we used
to lexicalise and display the dates of birth
to the crowdsourced speakers.
We used either of two formats at equal proportions:

\vspace{-0.5mm}
{\small
\begin{align*}
& (a)\ & \textbf{month=name} :& \ \textit{1 January|stycznia|janvier 2000} \\
& (b)\ & \textbf{month=number} :& \ \textit{1/1/2000}
\end{align*}}
The Sankey diagram\footnote{Sankey diagrams
    visualise the flow or route of communication (or other quantity) within a system to help locate the most important contributions to a flow. The width of the links between nodes 
    is proportional to the flow rate between them.} in Figure~\ref{fig:priming}~(top)
shows that $92\%$ of English speakers primed with the \textit{month=name} format
echoed this pattern in $\text{Q}_7$, and only $10\%$ of those switched to say the month's number in follow-up turns (similar results for pl-PL and fr-FR; see Appendix~\ref{appendix-dates} for their Sankey diagrams).
On the other hand, only $54\%$ of English speakers (cf. $26\%$ for pl-PL, $36\%$ for fr-FR; Appendix~\ref{appendix-dates}) primed with the \textit{month=number} format
echoed that pattern in $\text{Q}_7$,
and $77\%$ of those switched to say the month's name later.
Overall, the \textit{month=name} format (more lexical) had a stronger priming effect than the \textit{month=number} format (more symbolic),
and speakers say the month's name (more verbose) increasingly after reprompts ($\text{Q}_8$ and $\text{Q}_9$).

\begin{table}[t]
\begin{center}
\def\arraystretch{0.99}
{\small
\begin{tabular}{rrrrr}
\toprule
& & \multicolumn{3}{c}{\textbf{Locale}} \\
\cmidrule{3-5}
\multicolumn{2}{r}{\textbf{counts (unique)}}
& \textbf{en-GB} & \textbf{pl-PL} & \textbf{fr-FR}\\
\midrule
\multirow{6}{*}{\rotatebox[origin=c]{90}{\textbf{KB}}}
& \#profiles
& 10,000 & 10,000 & 10,000    \\
& \#postcodes 
& 2,000 & 2,000 & 2,000  \\
& \#names(first) 
& 364 & 153 & 216  \\
& \#names(last) 
& 500 & 3,455 & 400  \\
& \#names(full) 
& 9,412 & 9,923 & 9,433 \\
& \#DoBs 
& 8,884 & 8,862 & 8,862 \\
\midrule
\multirow{4}{*}{\rotatebox[origin=c]{90}{\textbf{Dialogues}}}
& \#dialogues 
& 1,407 & 1,991 & 2,108 \\
& \#turns
& 12,663 & 17,919 & 18,972 \\  
& \#speakers 
& 1,081 & 803 & 521 \\
& \#profiles 
& 886 & 961 & 1,464 \\
\bottomrule
\end{tabular}
}%
\end{center}
\caption{Size of the created EVI Knowledge Bases and the collected Conversational Dataset.}
\label{tab:describe_data}
\end{table}

\rparagraph{Diversifying Spoken Spellings}
Our primes to diversify the speakers' spelling choices
were the agent reprompts in the $\text{Q}_3$ 
that read back partial spellings of postcodes to the speaker.
We used either of two strategies at equal proportion:

\vspace{-0.5mm}
{\small
\begin{align*}
(a)\ \textbf{spell=naive}:& \ \textit{A B one two C D} \\
(b)\ \textbf{spell=nato}:\footnotemark& \ \textit{Alfa Bravo one two Charlie Delta}\vspace{-1mm}
\end{align*}}\footnotetext{The NATO phonetic alphabet substitutes a word for each letter to be easily understood in voice communications; \url{https://www.nato.int/cps/en/natohq/declassified_136216.htm}}These strategies acted as primes that \textit{entrained}
the speaker concerning their spelling strategy.

Figure~\ref{fig:priming}~(bottom) shows
that only $1\%$ of en-GB speakers
spontaneously used NATO spelling before/without encountering
the \textit{spell=nato} strategy in $\text{Q}_3$.
Conversely,
using the \textit{spell=nato} strategy
entrained ~52\% of speakers to adopt that strategy
in their response to $\text{Q}_3$.
Entrainment weakens over time:
only $28\%$ of entrained speakers
remained entrained by $\text{Q}_6$. 
Postcodes do not contain letters in the pl-PL and fr-FR locales, so both spelling strategies are equivalent.
Only $0.5\%$ of pl-PL and $0.1\%$ of fr-FR speakers spontaneously used complex spelling strategies (listed in Appendix~\ref{appendix-nato}).

{
In conclusion,
we validated that priming and entrainment 
are effective tools to subtly guide speaker behaviour
towards desired patterns.
It is by varying those primes
that we could increase the variability of speaker behaviours in our dataset.
}

\begin{figure}[t]
\centering
\includegraphics[width=0.89\columnwidth,trim={70 30 70 5},clip]{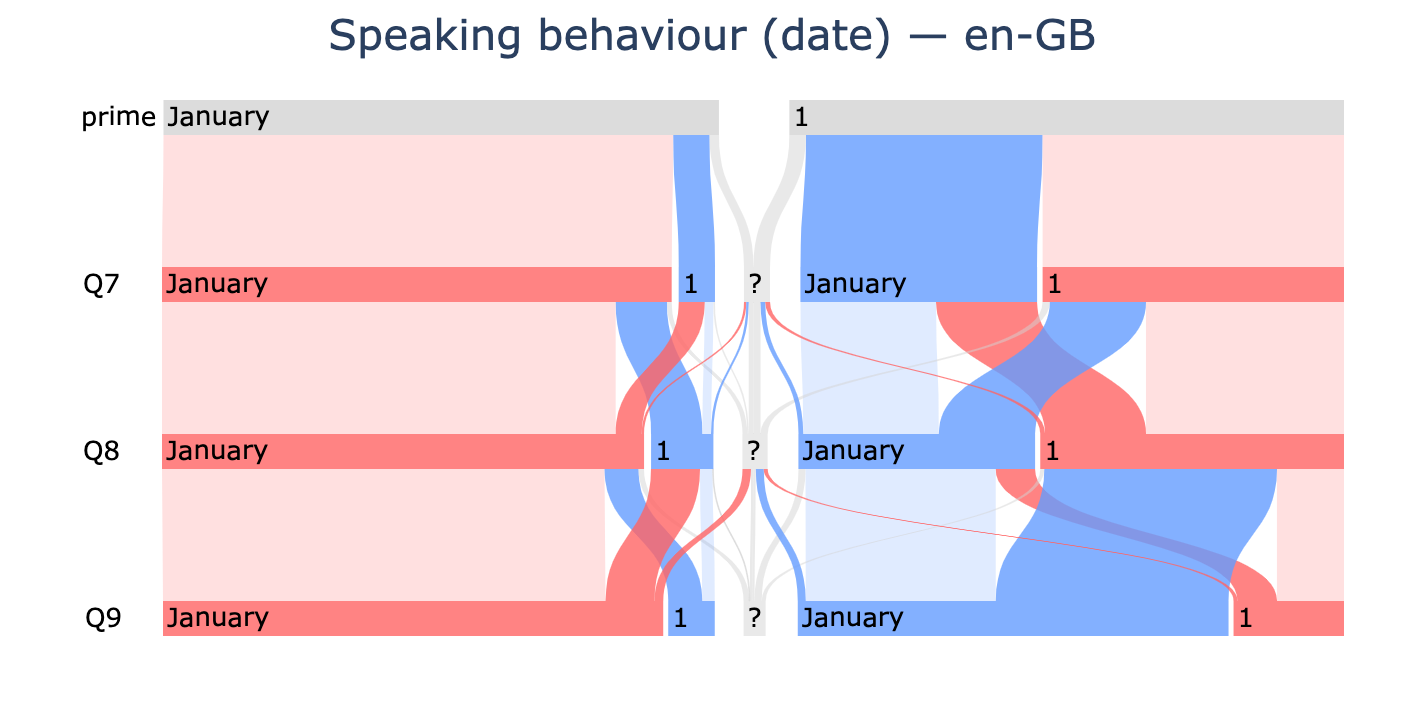}
\includegraphics[width=0.89\columnwidth,trim={50 30 70 5},clip]{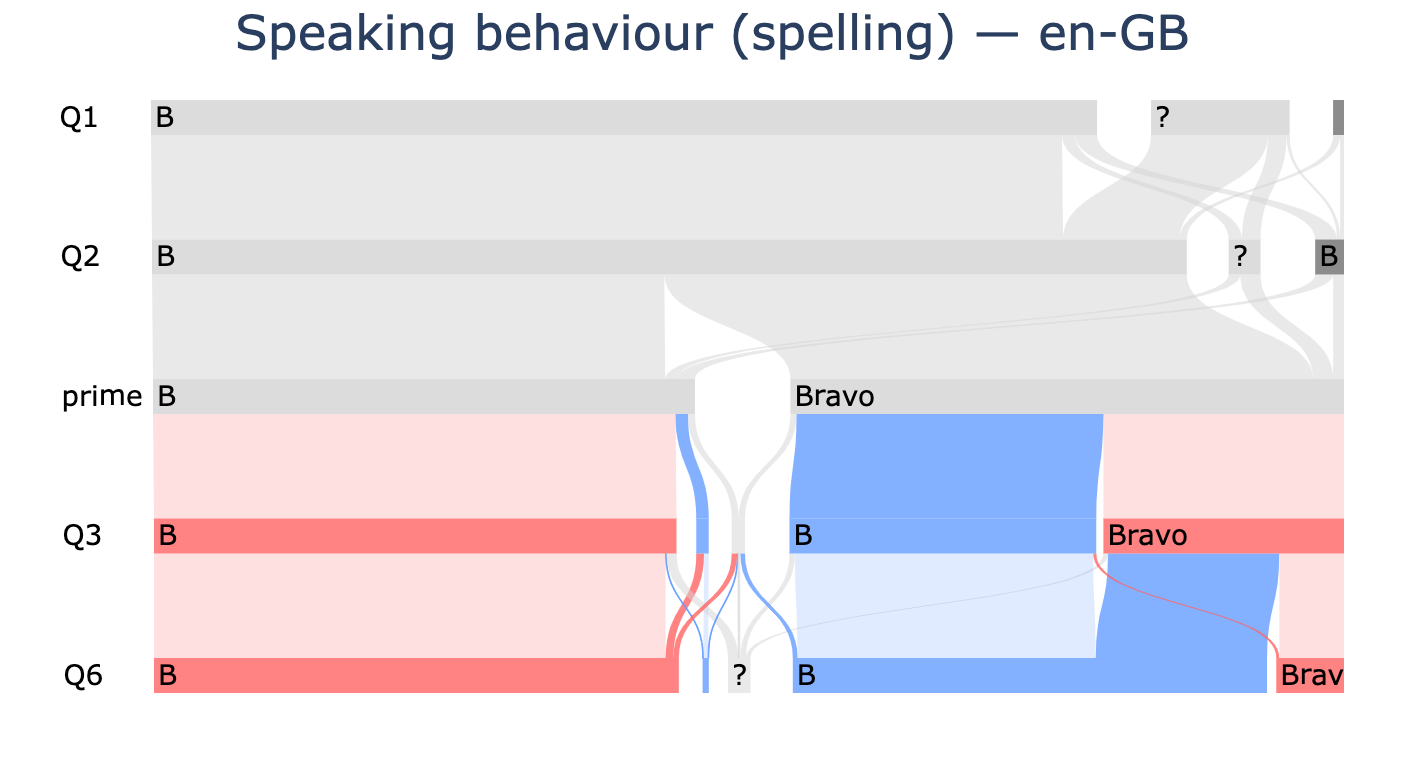}
\caption{
    Sankey diagrams that visualise
    priming and entrainment of speaker behaviour
    for dates (\textit{top}) and spelling (\textit{bottom}) for the British English locale.
    Transitions in the direction of priming in \textcolor{red}{red}; against, in \textcolor{blue}{blue}.
}
\label{fig:priming}
\end{figure}




\section{EVI-oriented Spoken Dialogue Systems}
\label{sec:methods}

This section presents the components of task-oriented spoken dialogue systems for EVI tasks and provides benchmark implementations for the upcoming experiments
(see Sections~\ref{sec:enrolment-experiments}, \ref{sec:verification-experiments}, and~\ref{sec:identification-experiments})


\subsection{Components of EVI Dialogue Systems}
\label{sec:evi-shared-components}

\sparagraph{Automatic Speech Recognition (ASR)}
When collecting the EVI dataset,
we used Google's locale-specific speech-to-text\footnote{\url{https://cloud.google.com/speech-to-text}}
in streaming mode
to derive n-best transcriptions
and to implement quality control (see Subsection~\ref{sec:data-collection}).
Consequently, this is the ASR used in all experiments.
{The length of the n-best lists was on average $4.85$, $2.60$, and $8.65$ for English, Polish, and French, respectively (see Appendix~\ref{appendix-asr})
and was capped at a maximum of 20 items.}

\rparagraph{Natural Language Understanding (NLU)}
For each item, we use an appropriate resource
to extract values from the whole ASR n-best list
into an NLU results n-best list.
In our experiments,
we first preprocess
to normalise numbers (`one'$\rightarrow$`1')
and letter spellings (`Bravo|[B~for~B.*]'$\rightarrow$`B'),
and then extract values
for postcodes
using
locale-dependent regular expressions
(`A(A)9(A)~9AA' for {en-GB}; `99999' for {pl-PL} and {fr-FR});
for names,
the lists of names from
the US Census\footnote{\url{ https://www.census.gov/topics/population/genealogy/data/1990_census/1990_census_namefiles.html}} and other sources~\cite{NameDataset2021};
and for {dates},
the \textit{dateparser} package.\footnote{\url{https://dateparser.readthedocs.io/} it is a python package that can parse localised dates in any string format}
Using these resources, we define two NLU models for value extraction:
the \texttt{cautious} model requires whole-string match,
whereas the \texttt{seeking} model searches for (potentially overlapping) substring matches.

\rparagraph{Top-Level Policy}
All EVI tasks share a common sequence of \textit{dialogue acts} (DAs):
the agent asks (\texttt{request} DA) the user
to input the value (\texttt{inform} DA) of each profile item successively,
with a limited number of re-prompts per item.
In the experiments,
the order of items is: postcode, full name, and date-of-birth,
with up to $3$ attempts per item
(fixed at the time of dataset collection; see Subsection~\ref{sec:data-collection}).

\rparagraph{Task-Level Dialogue Management}
Each of the three tasks requires task-specific \textit{dialogue state tracking} (DST)
and \textit{dialogue policy}.
The DST model tracks and updates the system's state and belief about the values of items and the candidate profiles,
whereas dialogue policy selects the following system action
(e.g. re-prompt user, proceed to next item, terminate task)
and interacts with the profiles KB.
We define the task-specific DST models and policies in more detail
in Subsections~\ref{sec:enrolment-policy}, \ref{sec:verification-policy}, and~\ref{sec:identification-policy}.

\rparagraph{Integration with the Profiles KB}
For enrolment,
the system needs write access to the KB to store the extracted profile; for identification, the system needs read access to the KB to retrieve candidate profiles via a dynamic sequence of queries;
and for verification,
the claimed profile in the KB is previously made available from an upstream identification process (cf. Fig.~\ref{fig:tasks}).
In the experiments,
we do not explicitly model KB integration for enrolment (write-only access) and verification (downstream of identification);
for identification, we model a read-only KB integration
that supports querying \textit{by postcode} (exact match)
and an \texttt{oracle} that always includes the postcode of the correct profile in the query, regardless of the NLU results.

\rparagraph{Natural Language Generation (NLG)}
When collecting the dataset,
we used \textit{scripted} prompts (Subsection~\ref{sec:data-collection})
translated for each locale (Appendix~\ref{sec:appendix-qs}).

\rparagraph{Text-to-Speech (TTS)}
We used Google's\footnote{\url{https://cloud.google.com/text-to-speech}}
locale-specific TTS when collecting the EVI dataset.


\subsection{Enrolment Models and Policies}
\label{sec:enrolment-policy}

\sparagraph{Enrolment DST and Model}
We track the value of each item, which is initially \textit{undefined}.
After each user input for an item,
we may use the NLU n-best results to update its value.
When the enrolment policy terminates,
the enrolment model straightforwardly builds 
the new profile from the tracked items.
In the experiments,
we update an item's value
with its latest \textit{top-1} result of the NLU (if not empty).

\rparagraph{Enrolment Policy}
The task-level policy
determines when to proceed to the next item,
and decides when to terminate enrolment.
The policy
(re)prompts the user about an item
until
either the DST returns a well-defined value 
or the top-level policy reaches the limit for attempts
($3$; see Subsection~\ref{sec:evi-shared-components}).
After exhausting all items, the policy terminates
and writes the new profile into the KB.


\subsection{Verification Models and Policies}
\label{sec:verification-policy}

\sparagraph{Verification DST and Model}
We track a verification score for each item $s_{item}$ as follows (cf. Eq.~\ref{eq:verify-profile}):

\vspace{-2mm}
{\small
\begin{equation}
\begin{gathered}
s_\textit{item} = 
\textbf{\textrm{score}}(\textrm{item}(p_\textit{claimed}), \textrm{item}(T_\textit{dialogue}))
\in [0,1],
\end{gathered}
\label{eq:item-score}
\end{equation}}%
The scores are initially undefined,
and we track their maximum evaluation after each user input.
For the experiments,
we define the following scoring models:
the \texttt{random} model samples from the $[0,1]$ uniform distribution;
the \texttt{exact} model
returns ~$1$ 
if the value from the claimed profile
exactly matches any NLU n-best result, else, $0$ (\textit{undefined} for no NLU results);
and the \texttt{fuzzy} model
returns the best \textit{fuzzy match} score
between the value from the claimed profile
and all NLU n-best results (\textit{undefined} for no NLU results).
We implement this as the normalised Levenshtein edit distance
using the Wagner–Fischer algorithm~\cite{wagner1974string}.
Finally, we evaluate a logical expression under \textit{fuzzy logic}
to combine all item-level scores (Eq.~\ref{eq:item-score})
into a profile-level score as follows (see Eq.~\ref{eq:verify-profile}):

{\small
\begin{equation}
\begin{gathered}
s_\textit{profile} =
s_\textit{postcode} \ \opAnd \ s_\textit{dob} \ \opAnd \\
(s_\textit{name\_full} \ \opOr (s_\textit{name\_first} \ \opAnd \ s_\textit{name\_last})) \\
\end{gathered}
\label{eq:profile-score}
\end{equation}}%
\textit{Fuzzy logic}~\cite{zadeh1996fuzzy}
is a many-valued logic wherein truth values are real numbers in $[0,1]$ that represent degrees of truthfulness and reasons using fuzzy logic operators (analogous to Boolean logic's AND, OR, and NOT).
In the experiments, we choose the standard fuzzy logic operators~\cite{zadeh1996fuzzy}:

{\small
\begin{equation}
\begin{aligned}
\textbf{Boolean}      & \longleftrightarrow \textbf{Fuzzy} \\
\opAnd (x,y) & \longleftrightarrow \min(x,y) \\
\opOr (x,y)  & \longleftrightarrow \max(x,y) \\
\opNot (x)   & \longleftrightarrow 1 - x
\end{aligned} \label{eq:logic-zadeh}
\end{equation}}%

\sparagraph{Verification Policy}
The task-level policy
determines when to proceed to the next item,
and decides when to terminate the verification process.
The policy
(re)prompts the user about an item
until
either the DST returns a well-defined score (Eq.~\ref{eq:item-score})
or the top-level policy reaches the limit for attempts
(again, $3$).
The policy terminates
either
after exhausting all items
or when it meets an \textit{early termination} criterion:
a low upper bound on the profile score
(i.e. Eq.~\ref{eq:profile-score} with $\textit{undefined} \equiv 1$ is below the verification threshold, $\theta$)
guarantees a negative verification outcome.
Upon termination, the policy returns the profile-level verification score (Eq. ~\ref{eq:profile-score} with $\textit{undefined} \equiv 0$).


\begin{table*}[t!]
\begin{center}
\def\arraystretch{0.97}
\resizebox{1.0\textwidth}{!}{
\begin{tabular}{cr rrrr rrrr rrrr rrrr}
\toprule
&
\multicolumn{1}{c}{\textbf{models}} &
\multicolumn{4}{c}{\textbf{Profile}} &
\multicolumn{4}{c}{\textbf{Postcode}} &
\multicolumn{4}{c}{\textbf{Name}} &
\multicolumn{4}{c}{\textbf{DoB}} \\
\cmidrule(lr){2-2}
\cmidrule(lr){3-6}
\cmidrule(lr){7-10}
\cmidrule(lr){11-14}
\cmidrule(lr){15-18}
&
\multicolumn{1}{c}{\textbf{nlu}} &
\textbf{P\%} & \textbf{R\%} & \textbf{F1\%} & \textbf{L} &
\textbf{P\%} & \textbf{R\%} & \textbf{F1\%} & \textbf{L} &
\textbf{P\%} & \textbf{R\%} & \textbf{F1\%} & \textbf{L} &
\textbf{P\%} & \textbf{R\%} & \textbf{F1\%} & \textbf{L} \\
\midrule
\multirow{2}{*}{\textbf{en-GB}}
& cautious
& \textbf{38.83} & \textbf{30.27} & \textbf{34.02} & 4.15
& \textbf{69.08} & \textbf{55.20} & \textbf{61.37} & 1.83
& \textbf{65.88} & \textbf{64.88} & \textbf{65.38} & 1.12
& \textbf{80.37} & \textbf{78.97} & \textbf{79.66} & 1.21 \\
& seeking
& 27.44 & 23.34 & 25.22 & \textbf{3.86}
& 59.90 & 51.16 & 55.18 & \textbf{1.70}
& 63.74 & 63.51 & 63.63 & \textbf{1.10}
& 63.86 & 63.58 & 63.72 & \textbf{1.07} \\
\midrule
\multirow{2}{*}{\textbf{pl-PL}}
& cautious
& \textbf{66.41} & \textbf{60.37} & \textbf{63.25} & 3.98
& \textbf{95.51} & \textbf{91.91} & \textbf{93.68} & 1.51
& \textbf{71.86} & \textbf{69.26} & \textbf{70.54} & \textbf{1.20}
& \textbf{92.92} & \textbf{90.31} & \textbf{91.59} & 1.26 \\
& seeking
& 53.07 & 51.63 & 52.34 & \textbf{3.69}
& 87.85 & 86.44 & 87.14 & \textbf{1.38}
& 69.76 & 69.16 & 69.46 & \textbf{1.20}
& 82.83 & 82.37 & 82.60 & \textbf{1.11} \\
\midrule
\multirow{2}{*}{\textbf{fr-FR}}
& cautious
& \textbf{34.22} & \textbf{30.37} & \textbf{32.19} & 3.85
& \textbf{77.62} & \textbf{72.09} & \textbf{74.75} & 1.50
& 44.21 & 44.00 & 44.10 & \textbf{1.06}
& \textbf{90.81} & \textbf{86.81} & \textbf{88.76} & 1.29 \\
& seeking
& 26.46 & 24.68 & 25.54 & \textbf{3.63}
& 75.03 & 70.43 & 72.66 & \textbf{1.46}
& \textbf{44.27} & \textbf{44.19} & \textbf{44.23} & \textbf{1.06}
& 72.12 & 71.57 & 71.84 & \textbf{1.10}\\
\bottomrule
\end{tabular}
}
\end{center}
\caption{
Results for enrolment task:
Precision (P), Recall (R), F1 score, and average number of turns (L) for exact match of the whole profile and each of its items (postcode, full name, and date of birth (DoB)).
}
\label{tab:results_enrolment}
\end{table*}

\subsection{Identification Models and Policies}
\label{sec:identification-policy}

\sparagraph{Identification DST and Model}
We track the NLU n-best results from all turns
and the candidate profiles retrieved from the KB.
Our identification process
is an \textit{anytime} algorithm~\cite{zilberstein1996using}
that ranks the thus-far retrieved profiles
by a score (Eq.~\ref{eq:profile-score}),
excluding profiles below an identification threshold, $\theta$.
Following the literature on \textit{fuzzy retrieval}~\cite{zadrozny2009fuzzy},
instead of the standard fuzzy operators (Eq.~\ref{eq:logic-zadeh}),
we use \texttt{p-norm} fuzzy operators~\cite{salton1983extended}:\footnote{
The expression is based on the $L^p$-norm,
$||x||_p:=\left( \sum_{i=1}^{n}|x_i| ^p \right)^{1/p}$,
and is related to the generalised (aka power or Hölder) means~\cite{bullen2013handbook}.}

\vspace{-3mm}
{\small
\begin{equation}
\begin{aligned}
\opAnd^p \left(  s_1, ..., s_n \right)
&= 1 - \left( \frac{1}{n} \sum_{i=1}^{n} \left| 1-s_i \right| ^p \right)^{1/p} \\
\opOr^p \left( s_1, ..., s_n \right)
 &= \left( \frac{1}{n} \sum_{i=1}^{n} |s_i| ^p \right)^{1/p} \\
\end{aligned}
\label{eq:logic-pnorm}
\end{equation}
}%
In the experiments,
we approximate Eq.~\ref{eq:logic-pnorm} by the \textit{infinity-one} linear combination~\cite{smith1990aspects}:

\vspace{-1.5mm}
{\small
\begin{equation}
\begin{aligned}
\opOr_{\alpha} &=\alpha \opOr^{\infty} + (1-\alpha) \opOr^{1}\\
&=\alpha \max +(1-\alpha) \textrm{mean} \\
\opAnd_{\alpha} &=\alpha \opAnd^{\infty} + (1-\alpha) \opAnd^{1}\\
&=\alpha \min + (1-\alpha) \textrm{mean} \\
\end{aligned}
\label{eq:logic-infinity-one}
\end{equation}
}%
Note that $\opAnd_{1}=\opAnd^{\infty}=\min$ and $\opOr_{1}= \opOr^{\infty} =\max$ are the standard fuzzy operators (Eq.~\ref{eq:logic-zadeh}).
Finally, an identification \texttt{oracle}
always retrieves the correct profile if it is among the tracked candidates (i.e. retrieved from the KB).

\rparagraph{Identification Policy}
The task-level policy
queries the KB to retrieve candidate profiles
(see Subsection~\ref{sec:evi-shared-components}),
determines when to proceed to the next item,
and decides when to terminate the identification process.
The policy queries the KB with the NLU n-best results,
and sends the retrieved profiles to the DST.
Similarly to verification,
the policy
(re)prompts the user about an item
until
either the DST returns a well-defined score (Eq.~\ref{eq:item-score})
or the top-level policy reaches the limit for attempts
(again, $3$).
The policy terminates
after having exhausted all items,
or when the anytime result of identification is an empty list
and the KB cannot be queried by any upcoming item.
Upon termination, 
the policy returns the ranked list of identified profiles.



\subsection{Evaluating the EVI Tasks}
\label{sec:evaluation}

\begin{table}[t]
\begin{center}
\def\arraystretch{0.83}
\resizebox{\columnwidth * 1}{!}{
\begin{tabular}{c l rrr rrr rrr}
\toprule
 & \multicolumn{1}{c}{\textbf{Turns}} &
\multicolumn{3}{c}{\textbf{Postcode}} &
\multicolumn{3}{c}{\textbf{Name}} &
\multicolumn{3}{c}{\textbf{DoB}} \\
\cmidrule(lr){3-5}
\cmidrule(lr){6-8}
\cmidrule(lr){9-11}
&  \multicolumn{1}{c}{(Subsection~\ref{sec:data-collection})} &
\textbf{P\%} & \textbf{R\%} & \textbf{F1\%} &
\textbf{P\%} & \textbf{R\%} & \textbf{F1\%} &
\textbf{P\%} & \textbf{R\%} & \textbf{F1\%} \\
\midrule
\parbox[t]{2mm}{\multirow{4}{*}{\rotatebox[origin=c]{90}{\textbf{en-GB}}}}
&
single($\text{Q}_i$), $i=1,4,7$
& 68.17 & 32.80 & 44.29
& \textbf{67.35} & 61.71 & 64.40
& 81.48 & 69.00 & 74.73 \\
&
single($\text{Q}_i$), $i=2,5,8$
& 73.27 & 39.02 & 50.92
& 65.47 & 56.72 & 60.78
& 79.64 & 66.98 & 72.76 \\
&
single($\text{Q}_i$), $i=3,6,9$
& \textbf{75.95} & 37.64 & 50.34
& 20.03 & 10.26 & 13.57
& \textbf{86.31} & 71.97 & 78.49 \\
&
multi ($\text{Q}_{1-9}$)
& 69.08 & \textbf{55.20} & \textbf{61.37}
& 65.88 & \textbf{64.88} & \textbf{65.38}
& 80.37 & \textbf{78.97} & \textbf{79.66} \\
\midrule
\parbox[t]{2mm}{\multirow{4}{*}{\rotatebox[origin=c]{90}{\textbf{pl-PL}}}}
&
single($\text{Q}_i$), $i=1,4,7$
& 95.95 & 58.26 & 72.50
& \textbf{74.11} & 62.98 & 68.10
& 93.69 & 76.04 & 83.95 \\
&
single($\text{Q}_i$), $i=2,5,8$
& 97.37 & 79.96 & 87.81
& 73.62 & 62.08 & 67.36
& 93.33 & 77.30 & 84.56 \\
&
single($\text{Q}_i$), $i=3,6,9$
& \textbf{97.53} & 85.33 & 91.03
& 21.95 & 6.68 & 10.24
& \textbf{93.80} & 81.27 & 87.08 \\
&
multi ($\text{Q}_{1-9}$)
& 95.51 & \textbf{91.91} & \textbf{93.68}
& 71.86 & \textbf{69.26} & \textbf{70.54}
& 92.92 & \textbf{90.31} & \textbf{91.59} \\
\midrule
\parbox[t]{2mm}{\multirow{4}{*}{\rotatebox[origin=c]{90}{\textbf{fr-FR}}}}
&
single($\text{Q}_i$), $i=1,4,7$
& 80.76 & 51.59 & 62.96
& \textbf{45.06} & 42.86 & 43.93
& 91.21 & 73.42 & 81.36 \\
&
single($\text{Q}_i$), $i=2,5,8$
& 82.48 & 65.02 & 72.72
& 41.44 & 39.72 & 40.56
& \textbf{92.91} & 74.61 & 82.76 \\
&
single($\text{Q}_i$), $i=3,6,9$
& \textbf{83.09} & 65.07 & 72.98
& 2.64 & 1.85 & 2.18
& 92.02 & 76.08 & 83.29 \\
&
multi ($\text{Q}_{1-9}$)
& 77.62 & \textbf{72.09} & \textbf{74.75}
& 44.21 & \textbf{44.00} & \textbf{44.10}
& 90.81 & \textbf{86.81} & \textbf{88.76} \\
\bottomrule
\end{tabular}
}
\end{center}
\caption{Results for single- vs multi-turn value extraction with \texttt{cautious} NLU:
Precision (P), Recall (R), F1 score per item (postcode, full name, and date of birth).
}
\label{tab:results_nlu}
\end{table}

\begin{table*}[t!]
\def\arraystretch{0.99}
\begin{center}
\resizebox{1\textwidth}{!}{
\begin{tabular}{cc ccc ccc ccc}
\toprule
\multicolumn{2}{c}{\textbf{models}} &
\multicolumn{3}{c}{\textbf{en-GB}} &
\multicolumn{3}{c}{\textbf{pl-PL}} &
\multicolumn{3}{c}{\textbf{fr-FR}} \\
\cmidrule(lr){1-2}
\cmidrule(lr){3-5}
\cmidrule(lr){6-8}
\cmidrule(lr){9-11}
\textbf{nlu} & \textbf{V-model}
& \textbf{EER\%} & \textbf{FRR\%} & \textbf{L}
& \textbf{EER\%} & \textbf{FRR\%} & \textbf{L}
& \textbf{EER\%} & \textbf{FRR\%} & \textbf{L} \\
\midrule
cautious & random
& 32.95 & 54.70 & 4.15 (2.85)
& 17.28 & 30.99 & 3.98 (2.67)
& 22.50 & 49.83 & 3.85 (2.38) \\
cautious & exact
& 28.22 & 56.42 & 4.15(2.78)
& 17.60 & 35.20 & 3.98 (2.59)
& 27.48 & 54.95 & 3.85 (2.30) \\
cautious & fuzzy
& 22.47 & 24.27 & 4.15 (3.09)
& 6.88 & 11.24 & 3.98 (2.76)
& 11.01 & 29.06 & 3.85 (2.57) \\
seeking & random
& 31.86 & 58.67 & 3.86 (2.59)
& 17.83 & 38.93 & 3.69 (2.37)
& 24.11 & 49.22 & 3.63 (2.30) \\
seeking & exact
& 30.89 & 61.77 & 3.86 (2.50)
& 21.15 & 42.29 & 3.69 (2.31)
& 25.87 & 51.73 & 3.63 (2.25) \\
seeking & fuzzy
& \textbf{11.27} & \textbf{21.06} & 3.86 (2.84)
& \textbf{4.27} & \textbf{10.56} & 3.69 (2.53)
& \textbf{9.11} & \textbf{18.73} & 3.63 (2.53) \\
\bottomrule
\end{tabular}
}
\end{center}
\caption{Results of verification task:
Equal Error Rate (EER),
False Rejection Rate (FRR) $@\text{FAR}=1/10,000$,
and average number of turns
(L;
in parentheses: with early termination
$@\text{FAR}=1/10,000$).
}
\label{tab:results_verification}
\end{table*}

\sparagraph{Evaluating Enrolment}
Suitable evaluation metrics come from the area of information extraction: \textit{precision} (P), \textit{recall} (R), and \textit{F1} score, at the profile level or per item.\footnote{
Enrolment outputs (new profiles) are stored in the KB and fed into I\&V downstream tasks (Fig.~\ref{fig:tasks});
evaluating interactions among tasks is outside the scope of this paper.}

\rparagraph{Evaluating Verification}
The relevant literature describes two basic metrics~\cite{el2012evaluation}:
\textit{False Rejection Rate} (FRR)
is the proportion of genuine users
that the system incorrectly rejects as impostors;
conversely,
\textit{False Acceptance Rate} (FAR)
is the proportion of impostors
that the system incorrectly accepts as genuine.
Lower FRR indicates more usable systems,
and lower FAR, more secure,
e.g. $\text{FRR}=1\%$ at $\text{FAR}=1/10~000$
means that 1\% of genuine users will fail verification
at the security level
that falsely accepts $1$ impostor per $10,000$ impostor attempts.
\textit{Equal Error Rate} (EER) is the error rate when $\text{FAR}=\text{FRR}$;
it is a popular evaluation metric when a security level is not a priori specified.
Finally, the \textit{Detection Error Trade-off} (DET) graph
plots FRR (y-axis) against FAR (x-axis)
for varying values of the verification threshold ($\theta$)
to visualise usability across a range of security levels~\cite{martin1997det}.
%

\rparagraph{Evaluating Identification}
We rely on the \textit{identiﬁcation rate at rank $r$} (IR@r)~\cite{el2012evaluation}:
the proportion of identification transactions by pre-enrolled users
in which the correct profile is among the top-$r$ retrieved by the system.
It is equivalent to the familiar \textit{recall at rank} metric from information retrieval \cite{manning2008ir}.


\section{Experiments and Results}
\label{sec:experiments}

This section evaluates benchmarks and empirically explores the unique challenges of each EVI task.


\rparagraph{Experimental Setup}
For all experiments,
we deterministically simulate 
ground truths and 
user inputs
from our EVI KB and dataset,
respectively 
(see Subections~\ref{sec:kb-generation} and~\ref{sec:data-collection}).
The implementations of
ASR, top-level policy, NLG, and TTS
were set at the time of data collection
and are common for all EVI tasks (see Subsection~\ref{sec:evi-shared-components}).
Subsection~\ref{sec:evaluation} describes
the evaluation metrics for each task.


\subsection{Enrolment Experiments}
\label{sec:enrolment-experiments}

We evaluate the enrolment policy with \texttt{cautious} or \texttt{seeking} NLU (see Subsection~\ref{sec:evi-shared-components}).

\rparagraph{Results} Table~\ref{tab:results_enrolment} shows the impact of NLU
on enrolment task accuracy (i.e. precision, recall, F1),
for the whole profile and per item,
and the average dialogue length.
For whole profiles and almost all items,
\texttt{cautious} NLU,
which is more conservative and extracts fewer values,
yields better accuracy than
\texttt{seeking} NLU,
which is more liberal and over-extracts values.
Notably,
extraction of French names and English postcodes (alphanumeric)
was less accurate than for other locales (digit-only postcodes).

\rparagraph{Further Analysis} Table~\ref{tab:results_nlu}
shows {per item the accuracy} (i.e. precision, recall, F1)
of single- and multi-turn value extraction with the \texttt{cautious} model.
Consistently,
recall with multi-turn extraction
is higher than
single-turn recall of any individual turn.
Conversely,
individual single-turns yield the highest precisions.
Across locales,
the relevant precisions of turns
is retained for postcodes ($\small\text{Q}_{3}>\text{Q}_{2}>\text{Q}_{1}$) and 
names ($\small\text{Q}_{4}>\text{Q}_{5}>\text{Q}_{6}$)
(cf. Section~\ref{sec:data-collection}).
In particular,
extraction of name spellings ($\text{Q}_{6}$) is distinctly poor; this barely affects multi-turn performance,
because, on average, the system collects names before $\text{Q}_{6}$ (Table~\ref{tab:results_enrolment}).



\subsection{Verification Experiments}
\label{sec:verification-experiments}

\begin{figure}[t]
\centering
\includegraphics[width=0.99\columnwidth]{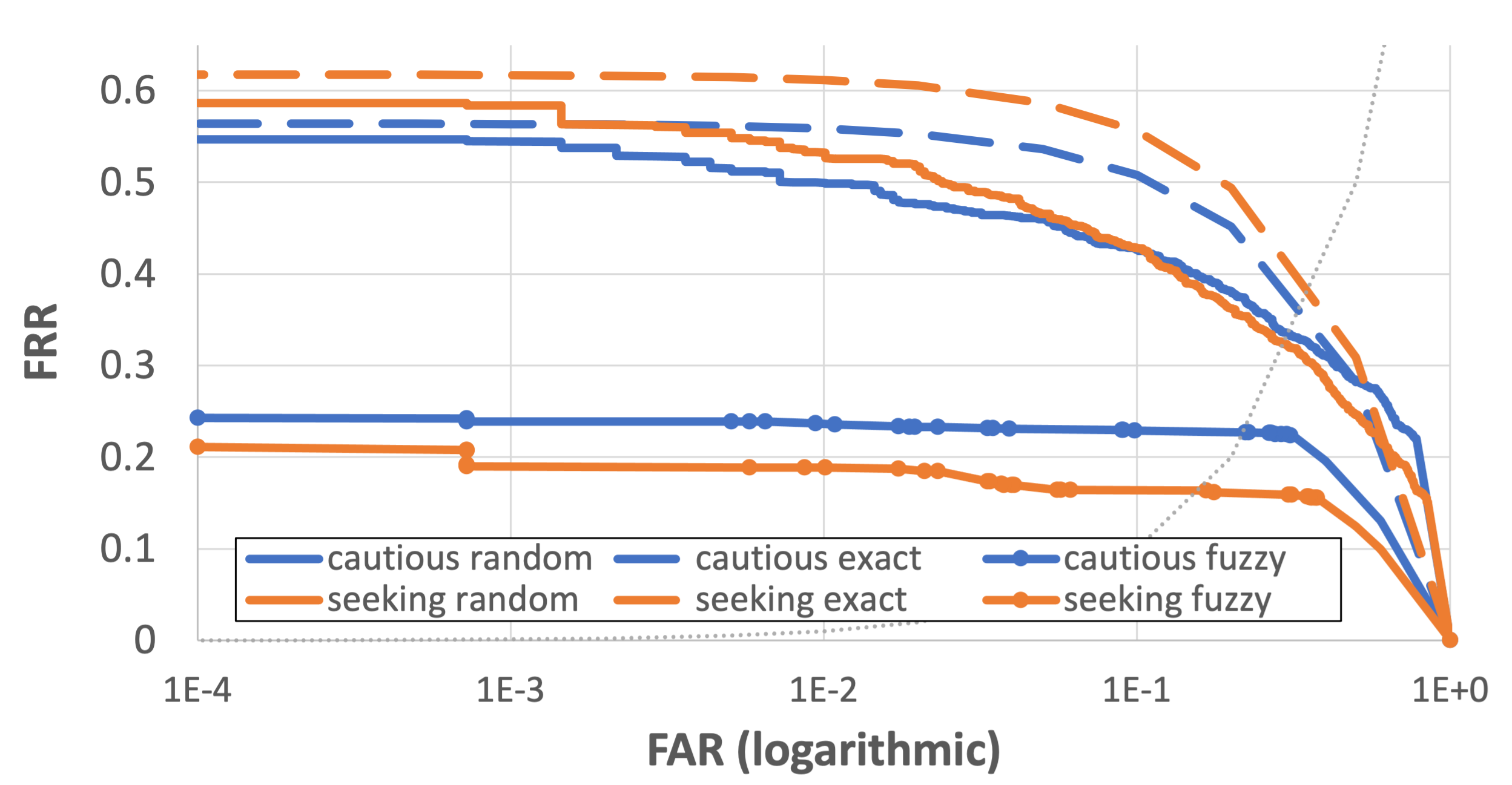}
\caption{
    Detection Error Trade-off (DET) curves
    for the en-GB locale.
    A curve that is closer to the bottom of the plot
    corresponds to better verification performance.
}
\label{fig:det}
\end{figure}

We evaluate the verification policy
with \texttt{cautious} or \texttt{seeking} NLU
and \texttt{random}, \texttt{exact}, or \texttt{fuzzy}
verification (Subsection~\ref{sec:verification-policy})
on the EVI dataset and KB (Section~\ref{sec:data}),
from which we sample genuine and impostor profiles
at a $1:1$ ratio.

\rparagraph{Results}
Table~\ref{tab:results_verification} shows
the impact of NLU and verification models
on the equal error rate (EER), the FRR at the $\text{FAR}=1/10~000$ security level and length.
Consistently,
\texttt{seeking} NLU with \texttt{fuzzy} verification
yields the best EER and FRR.
Interestingly,
\texttt{exact} verification
fails to improve reliably
over the \texttt{random} baseline.
Finally, early termination shortens verification length by 25-30\%.

\rparagraph{Further Analysis} Figure~\ref{fig:det}
shows the DET curves for the en-GB locale and all models.
\texttt{Exact} verification produces single points on the y-axis,
which we linearly interpolate to produce its DET curve.
Again,
\texttt{seeking} NLU with \texttt{fuzzy} verification
yields the best usability-security trade-off (lowest-lying curve)
for the whole range of security levels in the graph.
The same holds for the DET curves of the pl-PL and fr-FR
(shown in Appendix~\ref{appendix-det}).


\subsection{Identification Experiments}
\label{sec:identification-experiments}

We evaluate the identification policy
with \texttt{cautious} or \texttt{seeking} NLU (Subsection~\ref{sec:evi-shared-components}),
and no (\texttt{none}), \texttt{exact}, \texttt{fuzzy}, or \texttt{oracle} (upper bound) identification (Subsection~\ref{sec:identification-policy}).
We vary the $\alpha$ parameter
of the infinity-one \texttt{p-norm} (Eq.~\ref{eq:logic-pnorm}).

\rparagraph{Results} Table~\ref{tab:results_identification}
shows the impact of NLU and identification models on identification rate at rank 1 and identification length.
Without an explicit identification model (\texttt{none})
the agent cannot differentiate among multiple retrieved profiles
and accuracy is very low.
Consistently,
\texttt{seeking} NLU,
\texttt{fuzzy} models,
and $\alpha=0.5$ 
perform better than
\texttt{cautious} NLU,
\texttt{exact} matching,
and $\alpha=1$ (i.e. the standard fuzzy operators), respectively.
These effects are orthogonal:
\texttt{seeking} NLU with \texttt{fuzzy} model and $\alpha=0.5$
produces the best accuracy,
almost on par with the \texttt{oracle}.

\rparagraph{Further Analysis}
Most identification errors ($>98\%$) were caused by low recall:
the correct target profile was not included
in those returned by querying the KB with the NLU results, which is reminiscent of the \textit{unlinkable entity} (NIL) problem
from entity linking~\cite{ling2015design,hoffart2014discovering,mcnamee2009overview}.
Table~\ref{tab:results_kb_oracle} shows the upper bounds using a \texttt{KB oracle} (Subsection~\ref{sec:evi-shared-components}),
and corroborates the results of Table~\ref{tab:results_identification}.
The best combination (\texttt{seeking} NLU, \texttt{fuzzy} model and $\alpha=0.5$)
can achieve almost perfect performance as an upper bound.


\begin{table}[t]
\def\arraystretch{0.99}
\begin{center}
\resizebox{\columnwidth}{!}{
\begin{tabular}{cc ll ll ll}
\toprule

\multicolumn{2}{c}{\textbf{models}} &
\multicolumn{2}{c}{\textbf{en-GB}} &
\multicolumn{2}{c}{\textbf{pl-PL}} &
\multicolumn{2}{c}{\textbf{fr-FR}} \\
\cmidrule(lr){1-2}
\cmidrule(lr){3-4}
\cmidrule(lr){5-6}
\cmidrule(lr){7-8}
\textbf{nlu} & \textbf{I-model} &
\multicolumn{1}{c}{\textbf{IR@1}} & \multicolumn{1}{c}{\textbf{L}} &
\multicolumn{1}{c}{\textbf{IR@1}} & \multicolumn{1}{c}{\textbf{L}} &
\multicolumn{1}{c}{\textbf{IR@1}} & \multicolumn{1}{c}{\textbf{L}} \\
\midrule
cautious & none
& \ \ 9.90 & 3.64
& 19.74 & 3.86 
& 14.95 & 3.62 \\
seeking & none
& 10.04 & 3.54
& 19.89 & 3.71
& 15.09 & 3.46 \\
cautious & exact$(\alpha=1)$
& 50.22 & 3.64
& 65.90 & 3.86
& 48.50 & 3.62 \\
cautious & fuzzy$(\alpha=1)$
& 64.88 & 3.64
& 89.15 & 3.86
& 71.00 & 3.62 \\
seeking & exact$(\alpha=1)$
& 46.75 & 3.54
& 61.93 & 3.71
& 52.40 & 3.46 \\
seeking & fuzzy$(\alpha=1)$
& 66.18 & 3.54 
& 93.82 & 3.71
& 79.73 & 3.46 \\
cautious & exact$(\alpha=0.5)$
& 66.11 & 3.64
& 94.22 & 3.86
& 79.31 & 3.62 \\
cautious & fuzzy$(\alpha=0.5)$
& 66.33 & 3.64
& 94.32 & 3.86
& 78.97 & 3.62 \\
seeking & exact$(\alpha=0.5)$
& 67.27 & 3.54
& 94.88 & 3.71
& 80.35 & 3.46\\
seeking & fuzzy$(\alpha=0.5)$
& \textbf{67.77} & 3.54
& \textbf{95.13} & 3.71
& \textbf{80.83} & 3.46 \\
\midrule
cautious & \textit{oracle}
& 66.55 & 2.12
& 94.37 & 1.56
& 80.92 & 1.75 \\
seeking & \textit{oracle}
& 67.99 & 2.09
& 95.38 & 1.52
& 81.02 & 1.73 \\
\bottomrule
\end{tabular}
}
\end{center}
\caption{Results of identification task: Identification Rate at rank 1 (IR@1)
and average dialogue length (L).
}
\label{tab:results_identification}
\end{table}

\subsection{Directions for Further Research}
\vspace{-0.5mm}
Our findings highlight the most promising directions for further improvements.
In particular,
for enrolment: high-precision NLU and multi-turn belief tracking;
for verification: high-recall NLU and fuzzy matching;
and for identification: high-recall NLU, fuzzy retrieval, and boosting the recall of querying the KB.
All tasks can benefit from better multilingual NLU, and our dataset includes audios to encourage improvements in ASR.

\vspace{-0.5mm}

\begin{table}[t]
\def\arraystretch{0.87}
\begin{center}
\resizebox{\columnwidth}{!}{
\begin{tabular}{cc ll ll ll}
\toprule

\multicolumn{2}{c}{\textbf{models}} &
\multicolumn{2}{c}{\textbf{en-GB}} &
\multicolumn{2}{c}{\textbf{pl-PL}} &
\multicolumn{2}{c}{\textbf{fr-FR}} \\
\cmidrule(lr){1-2}
\cmidrule(lr){3-4}
\cmidrule(lr){5-6}
\cmidrule(lr){7-8}
\textbf{nlu} & \textbf{I-model} &
\multicolumn{1}{c}{\textbf{IR@1}} & \multicolumn{1}{c}{\textbf{L}} &
\multicolumn{1}{c}{\textbf{IR@1}} & \multicolumn{1}{c}{\textbf{L}} &
\multicolumn{1}{c}{\textbf{IR@1}} & \multicolumn{1}{c}{\textbf{L}} \\
\midrule
seeking & none
& 15.53 & 3.86
& 20.54 & 3.69
& 18.46 & 3.63 \\
seeking & exact$(\alpha=1)$
& 38.22 & 3.86
& 57.71 & 3.69
& 48.27 & 3.63 \\
seeking & fuzzy$(\alpha=1)$
& 81.86 & 3.86
& 95.63 & 3.69
& 90.18 & 3.63 \\
seeking & exact$(\alpha=0.5)$
& 96.60 & 3.86
& 97.79 & 3.69
& 97.63 & 3.63 \\
seeking & fuzzy$(\alpha=0.5)$
& \textbf{98.19} & 3.86
& \textbf{98.74} & 3.69
& \textbf{98.81} & 3.63 \\
\midrule
seeking & \textit{oracle}
& 100.00 & 1.00
& 100.00 & 1.00
& 100.00 & 1.00 \\
\bottomrule
\end{tabular}
}
\end{center}
\caption{Identification task with a \texttt{KB oracle}.}
\label{tab:results_kb_oracle}
\end{table}

\section{Related Work}

\vspace{-1mm}

\sparagraph{Authentication Tasks}
Our EVI tasks
seek to automate the process
of knowledge-based authentication~\cite{braz2006security,o2003comparing} in a voice communication context~\cite{o2006comparing,o2006speak,o2005query}
using task-oriented spoken dialogue systems.
We define and evaluate the tasks
analogously to 
automated systems for biometric authentication
(signatures,~\citealp{yeung2004svc2004};
fingerprints,~\citealp{maio2002fvc2000};
faces,~\citealp{phillips2003face};
irides,~\citealp{phillips2008iris};
and voice,~\citealp{doddington2000nist}).
\vspace{-0.5mm}

\rparagraph{Dialogues, NLP, and Logic}
Our {EVI benchmarks}
focus on speech recognition and spoken language understanding
of names~\cite{kaplan2020may,pappu2014knowledge},
dates~\cite{price2021hybrid},
and spellings~\cite{vertanen2012spelling,filisko2004error,chung2003automatic}.
Furthermore,
enrolment is
a particular case of the slot-filling dialogue task~\cite{young2002talking,bellegarda2014spoken};
and identification
is related to
information retrieval 
and shares challenges with entity linking~\cite{ling2015design,hoffart2014discovering,mcnamee2009overview}.
We extend fuzzy logic methods
from information retrieval~\cite{radecki1979fuzzy,zadrozny2009fuzzy,salton1983extended}
and
from multi-modal verification~\cite{lau2004fuzzy,conti2007fuzzy,azzini2007fuzzy}
to the context of spoken dialogues.

%

\rparagraph{Dialogue Datasets}
Research in dialogue systems
is driven by competitions~\cite{kim2019eighth,gunasekara2020overview}
and challenge datasets,
which may be
human-to-human
~\cite{schrading2015analysis,lowe2015ubuntu,ritter2010unsupervised},
machine-to-machine~\cite{shah2018building},
or human-to-machine (H2M) conversations;
about
single~\cite{coope2020span,wen2016network,hemphill1990atis}
or multiple domains~\cite{rastogi2020towards,zhu2020crosswoz,zang2020multiwoz,budzianowski2018multiwoz,asri2017frames};
in one or several languages~\cite{xu2020end,li2020mtop};
and
with written or spoken data~\cite{lugosch2019speech,li2018spoken,hemphill1990atis}.
Our EVI dataset is a spoken-language, multi-lingual, single-domain, human-to-machine challenge dataset for multiple tasks, which were not covered by any dialogue dataset from prior work.


\section{Conclusion}

We introduced novel spoken-dialogue tasks (knowledge-based enrolment, verification, and identification), the EVI multi-lingual spoken-dialogue dataset with $5,\!506$ dialogues,
and benchmark models, evaluations, and upper-performance bounds that leave ample margins for future improvements. 

\rparagraph{Limitations}
During data collection, our policy (fixed-length with reprompts for all items) might have caused artefacts in speaker behaviour (e.g. frustration, chuckling, simplification for later items). 
Additionally, speaker behaviour of crowd-sourced speakers
who impersonate a fake profile
will be qualitatively different to presenting one's own personal information
{(e.g. a young female speaker might be asked to impersonate an older male profile)};
however, ethical and privacy concerns preclude the publication of a dataset with real data.
Finally, our current evaluation considers each downstream task in isolation, although in practice they form a sequence (enrolment, identification, and then verification) that may propagate errors.

\rparagraph{Future Work} We invite the community to work on the novel EVI tasks and challenge dataset, which pose a variety of unresolved technical challenges:
speech recognition, multi-turn spoken language understanding, fuzzy matching and retrieval, etc.


\section*{Acknowledgements}
We are grateful to our colleagues at PolyAI for our many fruitful discussions.
We also thank the anonymous reviewers for their helpful suggestions.

\section*{Ethical Considerations}
\textcolor{black}{PolyAI} is ISO27k-certified and fully GDPR-compliant.
Before data collection,
we informed the crowd-sourced human workers that
their voluntary participation will allow us to
collect, store, publish, and use their fully-anonymous data for research purposes.
During data collection,
we did not ask workers for their own personal information (e.g. name, postcode);
instead, we provided fictional (but realistic looking) profiles for them to impersonate.
We instructed workers on how to hide their caller id,
we did not store any inbound phone numbers,
and we use fully anonymised identifiers in our dataset.
Finally,
we offered a fair compensation
(around the average hourly wage in the UK, pro-rata)
to all workers from all locales.

\bibliography{main}
%
\appendix

\renewcommand{\thesection}{\Alph{section}}

\section{Appendix}
\label{sec:appendix-qs}

This appendix
presents the scripted NLG prompts
(see Subsection~\ref{sec:data-collection} and Subsection~\ref{sec:evi-shared-components}).
For the British English locale (en-GB), see Subsection~\ref{sec:data-collection}.
All scripted prompts for the Polish locale (pl-PL):
\begin{quote}
{\small
\begin{enumerate}[label=Q\arabic*:,align=left, leftmargin=15pt]
    \item Podaj proszę swój kod pocztowy.
    \item Podaj go proszę jeszcze raz.
    \item Usłyszałam [1 2 3]. Podaj go jeszcze raz.
    \item Podaj teraz swoje imię i nazwisko?
    \item Podaj proszę swoję imię oraz nazwisko.
    \item Przepraszam, możesz przeliterować swoje imię i nazwisko?
    \item Jaka jest Twoja pełna data urodzenia?
    \item Podaj proszę datę urodzenia jeszcze raz.
    \item Usłyszałam [1 stycznia]. Podaj datę urodzenia jeszcze raz.
\end{enumerate}}
\end{quote}
All scripted prompts for the French locale \!(fr-FR):
\begin{quote}
{\small
\begin{enumerate}[label=Q\arabic*:,align=left, leftmargin=15pt, rightmargin=-10pt]
    \item Quel est votre code postal?
    \item Veuillez répéter votre code postal?.
    \item J'ai entendu [1 2 3]. Veuillez répéter votre code postal.
    \item Pourrais-je avoir votre nom et prénom?
    \item Pourrais-je avoir à nouveau votre nom et prénom
    \item Veuillez épeler votre nom complet?
    \item Quel est votre date de naissance?
    \item Pourrais-je avoir votre date de naissance.
    \item J'ai entendu [le 1er janvier]. Pourriez-vous répéter votre date de naissance.
\end{enumerate}}
\end{quote}

\section{Appendix}
\label{appendix-asr}

{
This appendix presents statistics of the ASR transcriptions
(see Subsections~\ref{sec:data-collection}~and~\ref{sec:evi-shared-components}).
In particular, the table shows the average length of the n-best lists
returned by the ASR per turn and for each locale.}

\begin{table}[!h]
\begin{center}
\def\arraystretch{0.95}
{\small
\begin{tabular}{rrrr}
\toprule
& \multicolumn{3}{c}{\textbf{Locale}} \\
\cmidrule{2-4}
\textbf{Turn} & \textbf{en-GB} & \textbf{pl-PL} & \textbf{fr-FR}\\
\midrule
1 
& 2.36 & 2.31 & 6.48\\
2
& 2.57 & 2.87 & 7.90\\
3
& 2.87 & 3.61 & 9.67\\
4
& 7.41 & 2.65 & 13.79\\
5
& 7.27 & 2.68 & 14.21\\
6
& 3.86 & 4.40 & 14.36\\
7
& 6.03 & 1.59 & 3.91\\
8
& 6.26 & 1.73 & 4.08\\
9
& 4.99 & 1.56 & 3.42\\
\midrule
all
& 4.85 & 2.60 & 8.65\\
\bottomrule
\end{tabular}
}%
\end{center}
\caption{{Average length of the ASR n-best lists in the EVI dataset. The maximum length is 20.}}
\end{table}

\section{Appendix}
\label{appendix-dates}

This appendix
presents Sankey diagrams for priming and speaker behaviour of dates (see Subsection~\ref{sec:data-analysis}). Transitions in the direction of priming in \textcolor{red}{red}; against, in \textcolor{blue}{blue}.
For the British English locale (en-GB), see Subsection~\ref{sec:data-analysis} and Fig.~\ref{fig:priming}.

\begin{figure}[h!]
\centering
\includegraphics[width=0.99\columnwidth]{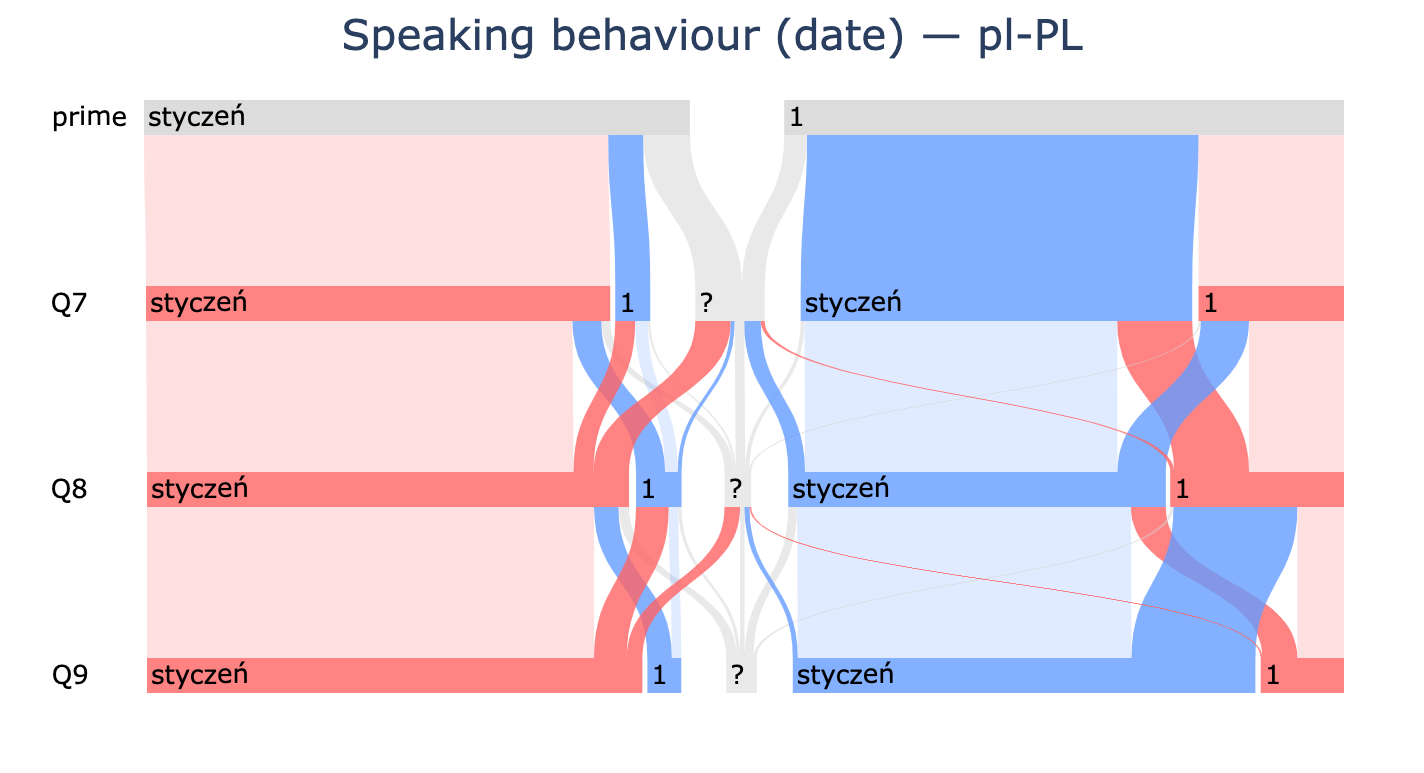}
\caption{Polish locale (pl-PL):
$85\%$ of speakers primed with \textit{month=name}
echoed this pattern in $\text{Q}_7$,
and only $10\%$ of those switched later;
$26\%$ primed with \textit{month=number} echoed and $71\%$ later switched.}
\end{figure}

\begin{figure}[h!]
\centering
\includegraphics[width=0.99\columnwidth]{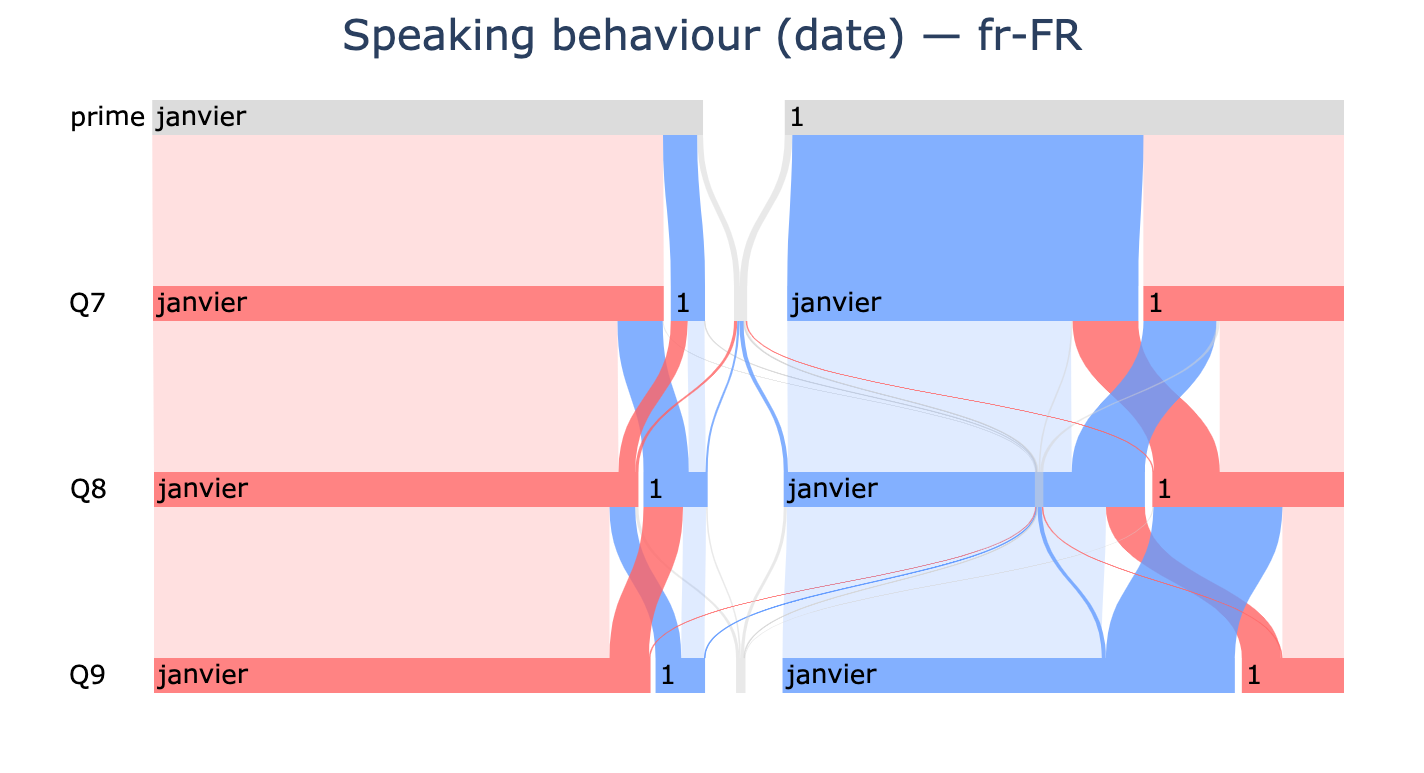}
\caption{
French locale (fr-FR):
$92\%$ of speakers primed with \textit{month=name}
echoed this pattern in $\text{Q}_7$,
and only $9\%$ of those switched later;
$36\%$ primed with \textit{month=number} echoed and $67\%$ later switched.}
\end{figure}

\section{Appendix}
\label{appendix-nato}

This appendix
presents the target names and top-1 ASR transcriptions
for all responses that employed complex spelling strategies.
For the British English locale (en-GB), consult the raw data (too many examples to list exhaustively).
All $10$ names with complex spelling transcriptions for the Polish locale (pl-PL):
\begin{quote}
{\small
\begin{itemize}[align=left,leftmargin=5pt,rightmargin=-10pt]
\item \textbf{[Juliusz Gwara]}: \textbf{J}oanna \textbf{U}rszula \textbf{L}idia \textbf{I}wona \textbf{U}rszula \textbf{S}abina Zenon \textbf{G}rażyna \textbf{W}aldemar \textbf{A}nna \textbf{R}oman \textbf{A}nna

\item \textbf{[Roksana Stypka]}: imię r jak \textbf{R}obert o jak \textbf{O}la kajak \textbf{K}atarzyna s jak \textbf{S}andra A jak \textbf{A}nna n jak \textbf{N}atalia a jak \textbf{A}nna nazwisko s jak \textbf{S}andra jak \textbf{T}adeusz \textbf{y} jak je t p jak \textbf{P}aulina k \textbf{K}atarzyna A jak \textbf{A}nna

\item \textbf{[Nela Domino]}: dobrze imię n jak \textbf{N}atalia e jak \textbf{E}lżbieta l jak \textbf{L}uiza A jak \textbf{A}nna nazwisko The jak \textbf{D}orota o jak \textbf{O}la i jak \textbf{I}rena n jak \textbf{N}atalia o jak \textbf{O}la

\item \textbf{[Róża Kochman]}: jak \textbf{r}yba \textbf{u z kreską} że jak \textbf{ż}aba A jak \textbf{A}nia

\item \textbf{[Ida Heinrich]}: i jak \textbf{i}gła d jak \textbf{D}anuta a jak \textbf{A}gnieszka ha jak \textbf{H}alina e jak \textbf{E}lżbieta I jak \textbf{i}gła n jak \textbf{N}atalia r jak \textbf{R}yszard i jak \textbf{i}gła c jak \textbf{c}ebula ha Jak \textbf{Ch}ełm

\item \textbf{[Sonia Dybiec]}: \textbf{S}abina \textbf{O}lga \textbf{N}atalia \textbf{I}rena \textbf{A}gnieszka \textbf{D}anuta \textbf{Y}eti \textbf{B}arbara \textbf{I}wona \textbf{E}lżbieta \textbf{C}elina

\item \textbf{[Kalina Hus]}: \textbf{K}rystyna \textbf{A}nna \textbf{L}ucyna \textbf{I}lona \textbf{N}atalia \textbf{A}nna \textbf{H}alina \textbf{U}rszula \textbf{S}abina

\item \textbf{[Elżbieta Minkina]}: \textbf{E}lżbieta \textbf{L}eokadia \textbf{Ż}aneta \textbf{B}olesław \textbf{I}lona \textbf{E}lżbieta \textbf{T}adeusz \textbf{A}nna \textbf{M}arlena \textbf{I}lona \textbf{N}atalia \textbf{K}arol \textbf{I}lona \textbf{N}atalia \textbf{A}nna

\item \textbf{[Justyna Grzelczyk]}: imię J Jak \textbf{J}ustyna u jak \textbf{U}rszula s jak \textbf{S}tefan te jak \textbf{T}eresa \textbf{y} jakie t n jak \textbf{N}atalia a jak \textbf{A}nna nazwisko g jak \textbf{G}rażyna r jak \textbf{R}obert \textbf{z} jak ze mną dieta l jak \textbf{L}uiza c jak \textbf{C}ezary \textbf{z} jak zenum \textbf{y} jakie t k jak \textbf{K}atarzyna

\item \textbf{[Piotr Kręcisz]}: p jak \textbf{p}ralka i jak \textbf{I}rena o jak \textbf{O}lga t jak \textbf{t}ata r jak \textbf{R}oman \textbf{k r a c z}

\end{itemize}}
\end{quote}
All $2$ names with complex spelling transcriptions for the French locale (fr-FR):
\begin{quote}
{\small
\begin{itemize}[align=left,leftmargin=5pt,rightmargin=-10pt]
\item \textbf{[Timothée Samson]}: est-ce qu'on sa vie à comme \textbf{A}lex matrix comme \textbf{S}ophie \textbf{O}livier comme \textbf{N}athalie

\item \textbf{[Constance Carlier]}: c'est con ce s'il a comme \textbf{A}lix elle comme elle est comme comme \textbf{É}milie el khomri
\end{itemize}}
\end{quote}
For the pl-PL and fr-FR locales, all listed examples are responses to $\text{Q}_6$ and arose spontaneously, without priming (see Subsection~\ref{sec:data-analysis}).

\section{Appendix}
\label{appendix-det}

This appendix
presents the DET plots (Subsection~\ref{sec:evaluation}) for the verification task experiments (Subsection~\ref{sec:verification-experiments}).
For the British English locale (en-GB), see Subsection~\ref{sec:verification-experiments} and Fig.~\ref{fig:det}.

\begin{figure}[h!]
\centering
\includegraphics[width=0.98\columnwidth]{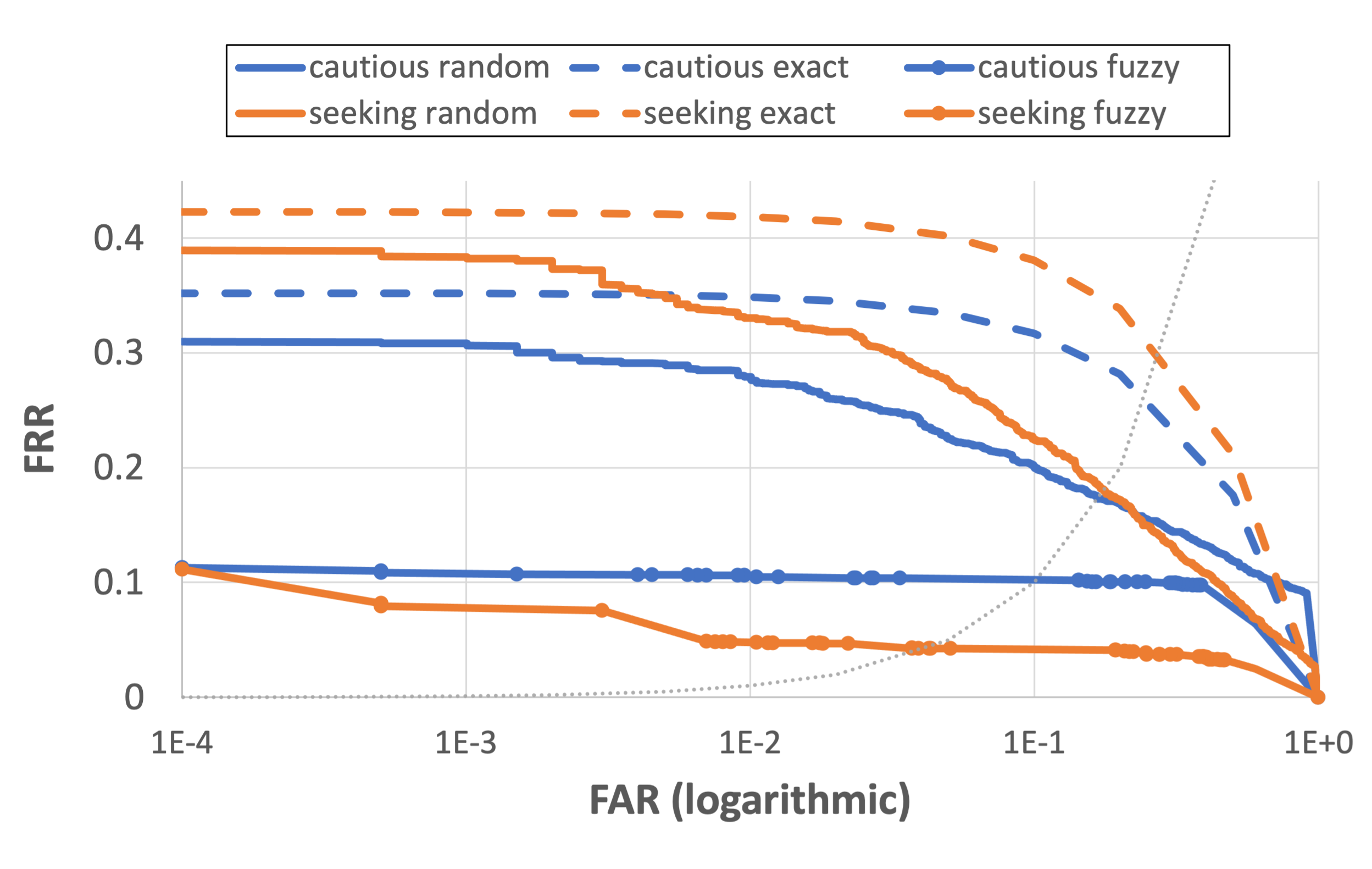}
\caption{DET curve for the Polish locale (pl-PL)}
\end{figure}

\begin{figure}[h!]
\centering
\includegraphics[width=0.98\columnwidth]{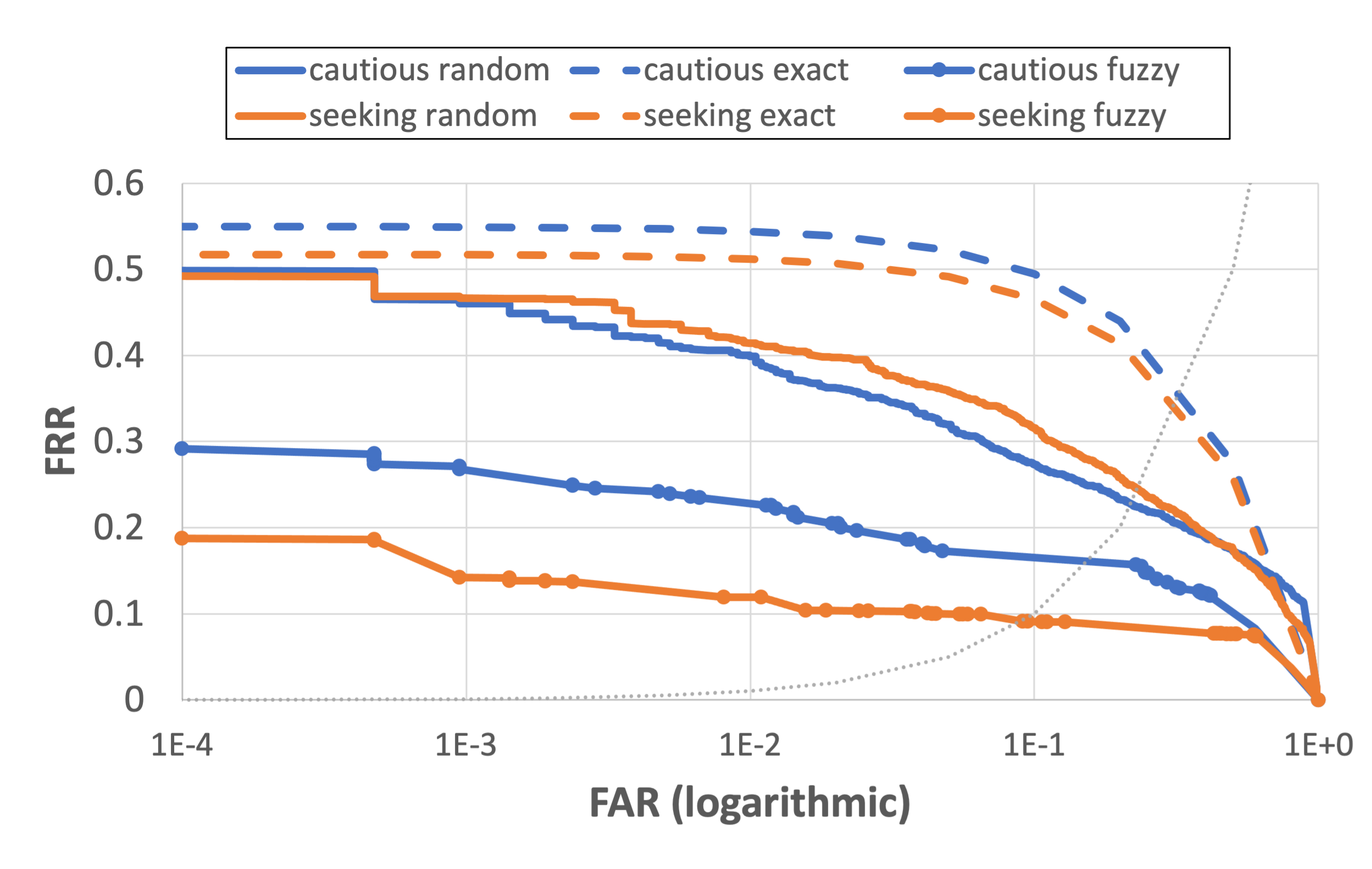}
\caption{DET curve for the French locale (fr-FR)}
\end{figure}

\end{document}